\documentclass[]{article}
\usepackage{amsmath, amssymb, amsthm,bbm,bm}

\RequirePackage[numbers]{natbib}%numbers
\usepackage{authblk} % for multiple authors/affiliations
\usepackage{orcidlink}

\usepackage{algorithm}
\usepackage{algpseudocode}

\usepackage[margin = 2.4cm]{geometry}
\newtheorem{remark}{Remark}
\newtheorem{theorem}{Theorem}

\newtheorem{proposition}{Proposition}
\newtheorem{example}{Example}
\newtheorem{assume}{Assumption}

\usepackage{setspace}
\title{\bf Sparse classification with positive-confidence data in high dimensions}

\author[1]{The Tien Mai$^{\orcidlink{0000-0002-3514-9636},}$}

\author[2]{Mai Anh Nguyen}

\author[3]{Trung Nghia Nguyen}

\affil[1]{\small
Norwegian Institute of Public Health, Oslo, 0456, Norway.
}
\affil[2]{\small
Seoul National University, South Korea.
}
\affil[3]{\small
Rutgers University, USA.
}

\date{\small
Email: \texttt{the.tien.mai@fhi.no}, 
}

\begin{document}
\maketitle

\begin{abstract}
High-dimensional learning problems, where the number of features exceeds the sample size, often require sparse regularization for effective prediction and variable selection. While established for fully supervised data, these techniques remain underexplored in weak-supervision settings such as Positive-Confidence (Pconf) classification. Pconf learning utilizes only positive samples equipped with confidence scores, thereby avoiding the need for negative data. However, existing Pconf methods are ill-suited for high-dimensional regimes. This paper proposes a novel sparse-penalization framework for high-dimensional Pconf classification. We introduce estimators using convex (Lasso) and non-convex (SCAD, MCP) penalties to address shrinkage bias and improve feature recovery. Theoretically, we establish estimation and prediction error bounds for the L1-regularized Pconf estimator, proving it achieves near minimax-optimal sparse recovery rates under Restricted Strong Convexity condition. To solve the resulting composite objective, we develop an efficient proximal gradient algorithm. Extensive simulations demonstrate that our proposed methods achieve predictive performance and variable selection accuracy comparable to fully supervised approaches, effectively bridging the gap between weak supervision and high-dimensional statistics.
\end{abstract}

Keywords: sparsity; SCAD; MCP; binary classification;  weak-supervised learning

\section{Introduction}

High-dimensional learning problems frequently arise in modern applications where the number of features can far exceed the sample size. In such regimes, achieving accurate prediction, stable parameter estimation, and reliable variable selection depends critically on the use of sparsity-inducing regularization \citep{hastie2009elements,buhlmann2011statistics,giraud2021introduction}. Classical approaches such as the Lasso \citep{tibshirani1996regression} have become standard tools, 
yet their behavior remains insufficiently understood in weak‐supervision settings, where only partial or indirect label information is available.

The shift toward weak supervision is driven by the fact that large, fully labeled datasets are often costly or fundamentally infeasible to obtain in fields like medical diagnosis, robotics, and bioinformatics. This has spurred a broad spectrum of research into paradigms such as semi-supervised learning \citep{sakai2017semi}, one-class classification \citep{khan2009survey}, and positive-unlabeled (PU) learning \citep{kiryo2017positive}. Collectively, these methods address the growing need for robust models that can operate effectively without ground-truth labels for every observation.

In this work, we focus on an increasingly relevant weak-supervision framework: positive-confidence (Pconf) classification \citep{ishida2018binary}. In Pconf learning, the training set consists exclusively of positive examples, each paired with a confidence score representing the probability that the sample belongs to the positive class. No negative samples are available.

Pconf classification offers distinct advantages over other weakly supervised methods. Unlike one-class classification, which is primarily descriptive and aimed at anomaly detection \citep{tax1999support}, Pconf is inherently discriminative; it seeks to separate classes even when one is unobserved. Furthermore, unlike PU learning, Pconf avoids the "notoriously challenging" problem of class-prior estimation \citep{christoffel2016class} because the confidence scores implicitly encode the necessary posterior structure. This allows for a clean Empirical Risk Minimization (ERM) formulation \citep{ishida2018binary}.

Despite its potential, current Pconf methodologies are designed for low-dimensional spaces \citep{ishida2018binary, li2024learning}. Naively applying ERM in high-dimensional settings leads to unstable estimates and poor generalization. While high-dimensional supervised classification is a mature field as in \citep{abramovich2018high,mai2023reduced,mai2024high,mai2025misclassification}, there is a significant gap in the literature regarding high-dimensional Pconf problems.

To address these challenges, in this paper, we propose a sparse-penalization framework for high-dimensional Pconf classification. 
We incorporate both convex (L1) and non-convex penalties, such as SCAD \citep{fan2001variable} and MCP \citep{zhang2010nearly}.
 While classical L1 regularization (Lasso) is popular due to its convexity and computational tractability \citep{tibshirani1996regression,mai2026heavylasso}, it is well known to suffer from shrinkage bias and often yields suboptimal variable selection when predictors are correlated. In contrast, non-convex penalties such as SCAD \citep{fan2001variable} and MCP \citep{zhang2010nearly} can mitigate these issues: they offer nearly unbiased estimation for large coefficients and improved feature recovery, making them particularly suitable for modern high-dimensional problems.

We provide rigorous theoretical analysis for the proposed Pconf estimator under an L1-penalized formulation. Building on the standard high-dimensional analysis for L1-regularized M-estimators \citep{negahban2012unified}, we establish estimation and prediction error bounds for convex-regularized Pconf learning. Under mild regularity conditions on the positive-confidence risk, 
we prove that Pconf learning achieves optimal sparse recovery rates comparable to standard supervised M-estimators, despite the total absence of negative labels. These results provide formal assurance that discriminative modeling with positive-confidence data remains statistically consistent and efficient even in high-dimensional regimes where the feature count exceeds the sample size.

To efficiently solve the penalized Pconf learning problem, we develop a proximal gradient algorithm tailored to the structure of the objective. The empirical Pconf risk is smooth and convex, while the sparsity penalty may be convex (L1) or nonconvex (SCAD, MCP). 
Proximal gradient methods \citep{parikh2014proximal} are ideally suited for such composite objectives because they cleanly separate the smooth loss from the nonsmooth penalty. Each iteration consists of a gradient step on the Pconf risk followed by a closed-form proximal update specific to the chosen penalty. For L1, this reduces to standard soft-thresholding, while for SCAD and MCP we employ their piecewise thresholding operators, which alleviate the bias of L1 and maintain computational simplicity. The resulting algorithm scales linearly with the sample dimension, requires no inner optimization loops, and remains stable under the nonconvex penalties. This provides an efficient and broadly applicable optimization framework for high-dimensional Pconf learning.

Through extensive simulations, we demonstrate that the proposed Pconf-regularized methods achieve predictive and estimation performance remarkably close to classical sparse classification methods that have access to fully labeled data. 
Despite relying only on positive-confidence labels, 
our Pconf-Lasso, Pconf-SCAD, and Pconf-MCP estimators attain comparable classification accuracy, low estimation error, and competitive variable-selection behavior. Across a range of sparse high-dimensional settings, the Pconf-based procedures recover the true active set with high true positive rates and controlled false discoveries, while requiring no negative labels. These results highlight that high-dimensional learning with positive-confidence data can be nearly as effective as standard supervised learning, confirming the practical value of our approach.

The remainder of the paper is organized as follows. Section \ref{sc_problem} introduces the standard classification framework and then formulates the classification problem under the positive confidence–only setting. Section \ref{sc_methoed_theory} presents the proposed methodology together with its theoretical guarantees. 
The proximal algorithm and its implementation details are described in Section \ref{sc_algorithm}. Simulation studies are reported in Section \ref{sc_simulations}, while an application to real data is provided in Section \ref{sc_realdata}. 
All technical proofs are deferred to the Appendix.
Our implementation is  available at \url{https://github.com/tienmt/Pconf_sparse}.

\section{Problem formulation}
\label{sc_problem}
\subsection{Classification problem}
Let $(\mathbf{x}, y)$ denote a pair consisting of a (high-dimensional) feature vector
$\mathbf{x} \in \mathbb{R}^d$ and its corresponding class label $y \in \{+1, -1\} $, 
assumed to follow an unknown joint distribution with density $p(\mathbf{x}, y)$. 
The goal of a general classification problem is to find a binary classifier $g(\mathbf{x}): \mathbb{R}^d \to \{1, -1\} $ that minimizes the misclassification risk:
\begin{equation}
\label{eq:classification_risk}
R(g) := \mathbb{E}_{p(\mathbf{x}, y)} \big[ \mathbf{1}(y \neq g(\mathbf{x})) \big],
\end{equation}
where $\mathbf{1}(\cdot)$ is the indicator function. Given an i.i.d. sample ${ (\mathbf{x}_i, y_i) }_{i=1}^n$ from $p(\mathbf{x}, y)$, the corresponding empirical risk is
\begin{equation}
\label{eq:classification_risk_empirical}
\hat{R}(g) := \frac{1}{n} \sum_{i=1}^n \mathbf{1}(y_i \neq g(\mathbf{x}_i)).
\end{equation}
However, the indicator function is non-smooth and non-convex, making direct minimization computationally intractable. A common solution is to replace the indicator with a convex surrogate loss $\ell$, yielding the expected loss
\begin{equation}
\label{eq:classification_risk_with_ell_loss}
R(g) := \mathbb{E}_{p(\mathbf{x}, y)} \big[ \ell(y , g(\mathbf{x})) \big],
\end{equation}
where $\ell(z)$ is a convex function and $\mathbb{E}_{p(\mathbf{x}, y)}$ denotes expectation over the unknown joint distribution. Since $p(\mathbf{x}, y)$ is unknown, empirical risk minimization (ERM) approximates this expectation using the average over the observed training data
$
n^{-1} \sum_{i=1}^n \ell(y_i , g(\mathbf{x}_i)).
$

\subsection{Classification with Positive-Confidence data}
In this work, we consider the Positive-Confidence (abbreviated as Pconf) problem, a special class of classification problems.
In Pconf settings, we aim to train a binary classifier only from positive data equipped with confidence, without negative data. 
More specifically, we are only given $ \{(\mathbf{x}_i, r_i)\}_{i=1}^n$,
where $\mathbf{x}_i$ is a positive pattern drawn independently from $p(\mathbf{x}|y=+1)$ and $r_i$ is the positive confidence given by $r_i = p(y=+1|\mathbf{x}_i)$. Since we have no access to negative data in the Pconf classification, we cannot directly employ the standard ERM approach.

Let $\pi_+=p(y=+1)$ and $ r(\mathbf{x}) = p(y=+1|\mathbf{x})$, and let $\mathbb{E}_+$ denote the expectation over $p(\mathbf{x}|y=+1)$. According to \cite[Theorem 1]{ishida2018binary}, the prediction error in Eq.~\eqref{eq:classification_risk_with_ell_loss} can be re-written as
\begin{equation}
\label{eq: rewritten_risk}
R(g) = \pi_+ \mathbb{E}_+ \left[\ell(g(\mathbf{x}))+\frac{1-r(\mathbf{x})}{r(\mathbf{x})}\ell(-g(\mathbf{x}))\right],
\end{equation}
given that $ r(\mathbf{x}) = p(y=+1|\mathbf{x})$ is strictly positive for all $\mathbf{x}$ sampled from $p(\mathbf{x})$.
Based on this result, \citet{ishida2018binary} had proposed an empirical risk minimization approach to tackle the problem of positive confidence classifiction. More specifically, for a linear classifier $ g(x) := \beta^T\mathbf{x} $, they studied the minimization of the following empirical risk 
\begin{equation}
\label{eq:_proposed_ERM}
\widehat{R}_n(\beta)=
\frac{1}{n} \sum_{i=1}^n\left[ \ell (\beta^T\mathbf{x}_i) 
+
\frac{1-r_i}{r_i} \ell ( - \beta^T\mathbf{x}_i) \right] 
.
\end{equation}

\begin{example}
 Consider a medical diagnosis task where the goal is to predict whether a patient has a particular disease \((y=+1)\) or not \((y=-1)\). In the Positive-Confidence (Pconf) setting, we only observe patients who are believed to be positive cases, but each comes with a confidence score reflecting diagnostic uncertainty. Let \(\mathbf{x}_i\) denote the feature vector of patient \(i\) (e.g., age, lab results, symptoms), and suppose a logistic regression model is used to estimate class probabilities,
 \(p_\beta(y=+1 \,|\, \mathbf{x}) = 1/ (1+ \exp(-\beta^\top \mathbf{x}) ) \).
 The confidence \(r_i\) associated with each observed positive sample is then \(r_i = p_\beta(y=+1 \,|\, \mathbf{x}_i)\), 
 which may be provided by a clinician or an existing probabilistic diagnostic system. For example, a patient with very strong symptoms might have \(r_i = 0.95\), while another with weaker or ambiguous symptoms might have \(r_i = 0.65\).
 Importantly, although all observed samples are drawn from \(p(\mathbf{x} \mid y=+1)\), their confidence values vary and reflect how likely each \(\mathbf{x}_i\) truly belongs to the positive class. Because no confirmed negative samples are available, standard empirical risk minimization for logistic regression cannot be applied directly, motivating specialized learning formulations for Pconf classification.

\end{example}

\begin{remark}
Note that a similar empirical formulation is as:
$
R_n^*(\beta)=
n^{-1} \sum_{i=1}^n 
[r_i \ell (\beta^T\mathbf{x}_i) 
+ (1-r_i)\ell ( -\beta^T\mathbf{x}_i) ] 
,
$
that reflects a different weighting strategy, where the positive loss receives weight $r_i$ and the negative loss receives weight $ 1-r_i $, making the formulation appear simple at first glance. However, as pointed out in \cite{ishida2018binary}, this empirical risk does not coincide with the classification risk defined in Eq.~\eqref{eq:classification_risk_with_ell_loss}. 
\end{remark}

\section{Sparse penalization approach for high-dimensional data}
\label{sc_methoed_theory}
\subsection{Proposed method}

We now propose a sparse penalization method to address positive-confidence (Pconf) classification in high-dimensional settings, where the number of features $d$ may exceed the number of observations $n$. Building on the confidence-weighted empirical risk formulation in \eqref{eq:_proposed_ERM}, we consider the following penalized objective:
\begin{equation}
\label{eq:proposed_ERM}
\hat{\beta} = \arg\min_{\beta \in \mathbb{R}^d}
\frac{1}{n} \sum_{i=1}^n
\left[ \ell (\beta^T \mathbf{x}_i)
+
\frac{1-r_i}{r_i} \ell ( - \beta^T \mathbf{x}_i) \right]
+
P_\lambda(\beta),
\end{equation}
where $\ell(\cdot)$ is a convex loss function (e.g., logistic or hinge loss), $r_i \in (0,1]$ is the confidence score for the $i$-th positive sample, and $P_\lambda(\beta)$ is a sparsity-inducing penalty term controlled by a tuning parameter $\lambda > 0$.

The addition of the sparsity penalty $P_\lambda(\beta)$ is crucial in high-dimensional settings to prevent overfitting, stabilize estimation, and perform variable selection. 
While sparsity is particularly critical when the dimensionality is large, it also plays a fundamental role in enabling variable selection more generally. In this study, we focus on three widely adopted sparsity penalties: the Lasso, the smoothly clipped absolute deviation (SCAD) penalty, and the minimax concave penalty (MCP).

\subsubsection{Lasso (L1) Penalty}
The Lasso penalty \citep{tibshirani1996regression} is defined as
\begin{equation}
P_\lambda^{\text{Lasso}}(\beta) = \lambda \sum_{j=1}^d |\beta_j|.
\end{equation}
Lasso encourages sparsity by shrinking small coefficients exactly to zero, thus performing automatic variable selection. Its convexity ensures that the optimization problem is tractable and globally solvable. However, Lasso is known to introduce shrinkage bias, particularly for large coefficients, and may select more variables than necessary when predictors are highly correlated.

\subsubsection{SCAD (Smoothly Clipped Absolute Deviation) Penalty}
The SCAD penalty \citep{fan2001variable} is a non-convex penalty designed to reduce bias in large coefficients while retaining sparsity. It is defined as $ P_\lambda(\beta) = \sum_{j=1}^d P_\lambda^{\text{SCAD}}(\beta_j), $ and

\begin{equation}
P_\lambda^{\text{SCAD}}(\beta_j) 
=
\begin{cases}
\lambda |\beta_j|, & |\beta_j| \le \lambda, 
\\
\dfrac{-\beta_j^2 + 2 a \lambda |\beta_j| - \lambda^2}{2(a-1)}, & \lambda < |\beta_j| \le a \lambda, 
\\
\dfrac{(a+1)\lambda^2}{2}, & |\beta_j| > a \lambda,
\end{cases}
\end{equation}
for some $a > 2$ (commonly $a=3.7$). SCAD preserves sparsity like Lasso but reduces bias for large coefficients, leading to better recovery of true signals in high-dimensional correlated data.

\subsubsection{MCP (Minimax Concave Penalty)}
The MCP \citep{zhang2010nearly} is another non-convex penalty with similar goals as SCAD. It is defined as $ P_\lambda(\beta) = \sum_{j=1}^d P_\lambda^{\text{MCP}}(\beta_j) $ and

\begin{equation}
P_\lambda^{\text{MCP}}(\beta_j) =
\begin{cases}
\lambda |\beta_j| - \dfrac{\beta_j^2}{2 \gamma}, & |\beta_j| \le \gamma \lambda, 
\\
\dfrac{\gamma \lambda^2}{2}, & |\beta_j| > \gamma \lambda,
\end{cases}
\end{equation}
where $\gamma > 0$ (often $\gamma = 3 $) controls the concavity. MCP penalizes small coefficients heavily to encourage sparsity but applies nearly zero penalty to large coefficients, mitigating bias while retaining good selection properties. Its concavity parameter $\gamma$ balances sparsity and estimation bias.

\subsection{Theoretical analysis}

Let $(\mathbf{x}_i, r_i)_{i=1}^n$ be i.i.d. samples, where $ \mathbf{x}_i $ are drawn from $p(\mathbf{x}|y=+1)$ and $r_i = p(y=+1|\mathbf{x}_i)$. Let the empirical Pconf risk be defined as
\begin{equation}
\label{eq:pconf_empirical_risk}
\widehat{R}_n(g)
:=
\frac{1}{n}\sum_{i=1}^n 
\left[
\ell(g(\mathbf{x}_i))
+
\frac{1-r_i}{r_i}\ell(-g(\mathbf{x}_i))
\right],
\end{equation}
where $\ell (\cdot) $ is a convex loss such as the logistic loss.
Consider the linear classifier $g_\beta(\mathbf{x}) = \mathbf{x}^\top \boldsymbol{\beta}$ and its L1-regularized empirical minimizer
\begin{equation}
\label{eq:L1_estimator}
\widehat{\boldsymbol{\beta}}
=
\arg\min_{\boldsymbol{\beta}\in\mathbb{R}^d}
\Big\{
\widehat{R}_n(\boldsymbol{\beta})
+
\lambda\|\boldsymbol{\beta}\|_1
\Big\}.
\end{equation}
Let $\boldsymbol{\beta}^*$ be the true minimizer of the population risk
\(
R(\boldsymbol{\beta})
=
\pi_+ \mathbb{E}_+ \big[
\ell(g_\beta(\mathbf{x})) + \frac{1-r(\mathbf{x})}{r(\mathbf{x})}\ell(-g_\beta(\mathbf{x}))
\big].
\)

\begin{assume}
\label{asume_main}
    Suppose the following assumptions hold:
\begin{enumerate}
    \item[(A1)] \textbf{Restricted Strong Convexity (RSC):} There exists $\kappa>0$ such that
    \[
    (\nabla \widehat{R}_n(\boldsymbol{\beta}) - \nabla \widehat{R}_n(\boldsymbol{\beta}^*))^\top (\boldsymbol{\beta} - \boldsymbol{\beta}^*)
    \geq
    \kappa \|\boldsymbol{\beta} - \boldsymbol{\beta}^*\|_2^2,
    \quad
    \forall \boldsymbol{\beta} \text{ in a neighborhood of } \boldsymbol{\beta}^*.
    \]
    \item[(A2)] \textbf{Gradient Regularity:} 
    The empirical gradient satisfies
    \(
    \|\nabla \widehat{R}_n(\boldsymbol{\beta}^*)\|_\infty
    \leq \lambda/2
    \)
    with probability at least $1-\delta$.
    \item[(A3)] \textbf{Sparsity:} The true parameter $\boldsymbol{\beta}^*$ is $s$-sparse, i.e., $\|\boldsymbol{\beta}^*\|_0 = s$.
\end{enumerate}
\end{assume}

Note that the restricted strong convexity (RSC) condition in Assumption A1 requires strong convexity of the objective function only in a neighborhood of the true parameter vector 
\( \boldsymbol{\beta}^* \). This type of localized convexity is standard in the theoretical analysis of regularized convex M-estimators \citep{negahban2012unified, agarwal2012noisy}. Assumption A2 imposes a gradient regularity condition, which is likewise standard; under additional assumptions on the design matrix, 
the loss function, and the weights \( (1 - r_i)/r_i \), this condition can be shown to hold. Finally, Assumption A3 posits that the true underlying model is sparse.

\begin{theorem}
\label{thm_pconf_estimation_error}

Suppose Assumption \ref{asume_main} hold.
Then for any $\delta\in(0,1)$, with probability at least $1 - \delta$, the L1-regularized estimator satisfies the following bounds:
\begin{align}
\label{eq:L1_l2_error}
\|\widehat{\boldsymbol{\beta}} - \boldsymbol{\beta}^*\|_2
&
\lesssim
\sqrt{s}\lambda, \\
\label{eq:L1_L1_error}
\|\widehat{\boldsymbol{\beta}} - \boldsymbol{\beta}^*\|_1
&
\lesssim
s\lambda.
\end{align}
Moreover, the excess risk obeys
\begin{equation}
\label{eq:excess_risk_bound}
R(\widehat{\boldsymbol{\beta}}) - R(\boldsymbol{\beta}^*)
\lesssim
s\lambda^2.
\end{equation}
\end{theorem}

The proof, given in Appendix \ref{sc_proof}, follows standard high-dimensional analysis for convex-penalized M-estimators \citep{negahban2012unified}, adapted to the Pconf setting. Under additional assumptions, we can explicitly verify Assumption A2 and provide explicit bounds for our results. This is presented in the following proposition.

\begin{proposition}[High-probability bound on gradient $\|\nabla \widehat R_n(\beta^*)\|_\infty$]
\label{lem:grad_concentration_bounded}

Assume that for all $i,j$ we have $\; |x_{ij}|\le B$ 
and $ \; |\ell'(t)|\le L_1$ for all $t$, and that the weights satisfy $\; \frac{1-r_i}{r_i}\le W$ almost surely. Define
$ \;
C_\psi := L_1(1+W).$
Then for any $\delta\in(0,1)$, with probability at least $1-\delta$,
\begin{equation}
\label{eq:grad_inf_bound_bounded}
\|\nabla \widehat R_n(\beta^*)\|_\infty
\le
C_\psi B \sqrt{\frac{\log(2d/\delta)}{2n}}.
\end{equation}
Consequently, the choice
\begin{equation}
\label{eq:lambda_choice_bounded}
\lambda 
=
C_\psi B \sqrt{\frac{2\log(2d/\delta)}{n}}
\end{equation}
ensures $\lambda \ge 2\|\nabla \widehat R_n(\beta^*)\|_\infty$ with probability at least $1-\delta$.
\end{proposition}

The results in Proposition \ref{lem:grad_concentration_bounded} say that under certain assumptions, the Assumption A2 above is satisfied. 
Moreover, it implies that there exist some choice of the tuning parameter $ \lambda \asymp \sqrt{\frac{\log (d)}{n}}  $,
so that we have up to some positive constants that
$$
\|\widehat{\boldsymbol{\beta}} - \boldsymbol{\beta}^*\|_2
\lesssim
\sqrt{\frac{s \log (d)}{n}},
\quad
\|\widehat{\boldsymbol{\beta}} - \boldsymbol{\beta}^*\|_1
\lesssim
s\sqrt{\frac{ \log (d)}{n}},
\quad
R(\widehat{\boldsymbol{\beta}}) - R(\boldsymbol{\beta}^*)
\lesssim
\frac{s \log (d)}{n}
.
$$
These rates nearly match the optimal rate for supervised classification as in \cite{abramovich2018high}.
We observe that the assumption of a uniform upper bound \( \frac{1-r_i}{r_i} \le W \) is equivalent to requiring that the confidence scores satisfy \( r_i \ge r_{\min} > 0 \). This positivity condition is natural in our setting, as it ensures that the Pconf weights remain well defined and uniformly bounded.

\section{Proximal gradient algorithm}
\label{sc_algorithm}
To optimize the penalized positive-confidence (Pconf) with logistic loss, we employ a \textit{proximal gradient method} \citep{parikh2014proximal} (also known as forward–backward splitting in the more general setting). 
This approach is particularly suited for our learning problems because it decouples the smooth empirical risk from the nonsmooth or nonconvex sparsity penalty.

For the optimization problem in \eqref{eq:proposed_ERM}, the empirical Pconf risk $\widehat{r}$ is smooth and convex. 
The penalty $P_\lambda (\mathbf{\beta})$ may be convex (e.g., $L_1$) or nonconvex (e.g., SCAD, MCP). Proximal gradient methods are appropriate here as they do not require convexity of the penalty and naturally handle non-smoothness.

\subsection{Smooth component: Gradient derivation}

Let $\ell(z) = \log(1+\exp(-z))$ be the logistic loss and $\sigma(z) = (1+\exp(-z))^{-1}$ be the sigmoid function, with derivative $\ell'(z) = -\sigma(-z)$.
For the Pconf loss term $ \,
\ell(g_i)+\alpha_i\ell(-g_i)$, the derivative with respect to the margin $g_i$ is:
\begin{align}
    d_i :=
    \frac{\partial}{\partial g_i} \left[ \ell(g_i)+\alpha_i\ell(-g_i) \right] 
    =
    -\sigma(-g_i) + \alpha_i\sigma(g_i).
\end{align}
Consequently, the gradient of the empirical risk is given by:
\begin{equation}
    \nabla_\mathbf{\beta} \widehat{R}_n (\mathbf{\beta}) = \frac{1}{n} \mathbf{X}^\top \mathbf{d},
,
\end{equation}
where $\mathbf{d} = (d_1,\ldots,d_n)^\top$.
%To prevent numerical overflow for large $|g_i|$, both the logistic loss and sigmoid functions are implemented in piecewise numerically stable forms using $\texttt{log1p}$ and equivalent formulations.

\subsection{Proximal Gradient Scheme}
Our algorithm is summarized in Algorithm \ref{algorithm_main}.
The proximal gradient update at iteration $t$ consists of two distinct steps:

{\bf 1. Forward (Gradient) Step.}
A gradient descent step is taken on the smooth Pconf loss with step size $\eta_t$:
\begin{align}
    \tilde{\mathbf{\beta}}^{(t+1)} &= \mathbf{\beta}^{(t)} - \eta_t \nabla_\mathbf{\beta} \widehat{R}_n (\mathbf{\beta}^{(t)}),
    .
\end{align}
In practice, a small fixed $\eta_t$ works efficiently due to the smoothness of the loss. This step handles the data-fitting component but ignores the sparsity constraint.

{\bf 2. Backward (Proximal) Step.}
The backward step applies the proximal operator of the penalty $P_\lambda (\boldsymbol{\beta})$. For the updated estimate $\tilde{\boldsymbol{\beta}}^{(t+1)}$ from the gradient step, the proximal update is defined as:
\begin{equation}
    \boldsymbol{\beta}^{(t+1)} 
    = \operatorname{prox}_{\eta_t, P_\lambda} \left(\tilde{\boldsymbol{\beta}}^{(t+1)} \right)
    = \operatorname*{arg\,min}_{\mathbf{u}\in\mathbb{R}^p} \left( \frac{1}{2\eta_t}\|\mathbf{u} - \tilde{\boldsymbol{\beta}}^{(t+1)}\|_2^2 
    +
    P_\lambda (\mathbf{u}) \right).
\end{equation}
This step acts component-wise on the vector $\tilde{\boldsymbol{\beta}}^{(t+1)}$. Let $z = \tilde{\beta}_j^{(t+1)}$ denote the $j$-th component before thresholding. Different choices of the penalty function $P_\lambda$ yield distinct closed-form proximal maps.

\begin{itemize}
    \item \textbf{L1 (Lasso):} The $L_1$ penalty induces the standard soft-thresholding operator:
    \begin{equation}
      {\beta}^{(t+1)}_j = \operatorname{sign}(z) \max\left(|z| - \eta_t\lambda, 0\right).
    \end{equation}
    This operator shifts the coefficient towards zero by $\eta_t\lambda$ and sets it to zero if the magnitude is within the noise level.

    \item \textbf{SCAD:}
    To mitigate the bias introduced by the Lasso, the SCAD penalty keeps large coefficients unpenalized.  The proximal operator is defined piecewise based on the magnitude of $z$. With hyperparameters $a > 2$ (typically $a=3.7$) and step size $\eta_t$:
    \begin{equation}
        {\beta}^{(t+1)}_j 
        = 
        \begin{cases} 
        \operatorname{sign}(z) \max(|z| - \eta_t\lambda, 0) & \text{if } |z| \le \lambda(1+\eta_t), \\
        \dfrac{(a-1)z - \operatorname{sign}(z)a\lambda\eta_t}{a - 1 - \eta_t} & \text{if } \lambda(1+\eta_t) < |z| \le a\lambda, \\
        z & \text{if } |z| > a\lambda.
        \end{cases}
    \end{equation}
    This operator transitions from soft-thresholding (Region 1) to linear interpolation (Region 2), and finally to the identity mapping (Region 3), thereby avoiding excessive shrinkage of strong signals.

    \item \textbf{MCP:}
    The MCP provides a faster transition to unbiasedness than SCAD. Its derivative is given by $P_\lambda'(\beta) = \lambda(1 - |\beta|/(a\lambda))_+ \operatorname{sign}(\beta)$. Assuming the step size satisfies $\eta_t < a$ (to maintain convexity of the subproblem), the proximal map is:
    \begin{equation}
        {\beta}^{(t+1)}_j = \begin{cases} 
        \dfrac{\operatorname{sign}(z) \max(|z| - \eta_t\lambda, 0)}{1 - \eta_t/a} & \text{if } |z| \le a\lambda, \\
        z & \text{if } |z| > a\lambda.
        \end{cases}
    \end{equation}
    Here, coefficients in the sparsity zone are soft-thresholded and then rescaled (inflated) by $(1 - \eta_t/a)^{-1}$ to offset the shrinkage bias, while coefficients larger than $a\lambda$ (typically $a=3$) are left unchanged.
\end{itemize}

The objective decomposition $F(\mathbf{\beta}) = \widehat{R}_n (\mathbf{\beta}) +  P_\lambda(\mathbf{\beta})$ 
allows the smooth part to be handled via efficient gradient evaluation and the nonsmooth part via a closed-form proximal operator. This avoids inner optimization loops, resulting in a computational cost of $\mathcal{O}(nd)$ per iteration, dominated by matrix–vector multiplication.

Proximal operators for SCAD and MCP do not require solving subproblems via coordinate descent. Their closed-form solutions ensure significantly reduced computational cost and guaranteed sparsity induction. Despite the nonconvexity of the full objective, proximal gradient methods converge to a critical point under standard assumptions (e.g., Lipschitz gradients).

\begin{algorithm}[t]
\caption{Proximal Gradient for sparse Positive-Confidence classification}
\begin{algorithmic}[1]
\State \textbf{Input:} Data $\{(\mathbf{x}_i, r_i)\}_{i=1}^n$, step size $\eta$, regularization $\lambda$, epochs $T$.
\State \textbf{Initialize:} $\mathbf{\beta}^{(0)} \sim \mathcal{N}(0, \sigma^2 \mathbf{I}_p)$,  $\alpha_i = (1-r_i)/r_i$.
\For{$t = 0, \ldots, T-1$}
    \State \textbf{1. Compute:}
  $\mathbf{g}^{(t)} = \mathbf{X} \mathbf{\beta}^{(t)}$
    \State \textbf{2. Gradient multipliers:}
    \State \quad $d_i^{(t)} = -\sigma(-g_i^{(t)}) + \alpha_i\sigma(g_i^{(t)})$
    \State \textbf{3. Gradient step:}
    \State \quad $\tilde{\mathbf{\beta}}^{(t+1)} = \mathbf{\beta}^{(t)} - \frac{\eta}{n} \mathbf{X}^\top \mathbf{d}^{(t)}$
    \State \textbf{4. Proximal update:}
    \State \quad $\mathbf{\beta}^{(t+1)} = \operatorname{prox}_{\eta\lambda P}(\tilde{\mathbf{\beta}}^{(t+1)})$
\EndFor
\State \textbf{Output:} Estimated classifier $(\hat{\mathbf{\beta}}) = (\mathbf{\beta}^{(T)})$.
\end{algorithmic}
\label{algorithm_main}
\end{algorithm}

\subsection{Selection of the regularization parameter $\lambda $}

The regularization parameter \(\lambda >0\) is selected by minimizing the cross-validated Pconf risk.
For each candidate value \(\lambda\), we perform \(K\)-fold cross-validation on the Pconf logistic risk:
\[
\widehat{R}(\beta)
=
\frac{1}{n}
\sum_{i=1}^n
\Big[
\ell(g_i) + \alpha_i \ell(-g_i)
\Big],
\quad
g_i = \beta^\top x_i,
\quad
\alpha_i = \frac{1-r_i}{r_i},
\]
where \(\ell(z)\) is the logistic loss.
In each fold, the model is trained on \(K-1\) subsets using the proximal gradient algorithm and evaluated on the held-out fold using the same risk \(\widehat{R}\). The cross-validated risk for a given \(\lambda\) is computed as
\[
\mathrm{CV}(\lambda)
=
\frac{1}{K}
\sum_{k=1}^K
\widehat{R}_{\mathcal{V}_k}(\lambda),
\]
where \(\widehat{R}_{\mathcal{V}_k}(\lambda)\) is the Pconf risk on the \(k\)-th validation set $ \mathcal{V}_k $.
The optimal regularization level is chosen as
\[
\lambda_{\mathrm{opt}}
= \arg\min_{\lambda \in \Lambda}
\mathrm{CV}(\lambda),
\]
that is, the value that minimizes the estimated out-of-sample Pconf risk. This ensures that the selection criterion is fully aligned with the objective optimized during training, and avoids dependence on surrogate performance measures such as accuracy (for which we do not have training labels).

Our codes is  available at \url{https://github.com/tienmt/Pconf_sparse}.

\section{Simulations}
\label{sc_simulations}
\subsection{Data generation and setup}
We study a high-dimensional binary classification problem in which the training data consist of feature vectors together with positive-confidence (Pconf) information, while full class labels are used only for a baseline comparison method (e.g., GLMNET). The aim is to evaluate classifiers that rely solely on confidence scores for observed positive samples.

We generate a training dataset with \(n\) observations and \(d\) features. The true underlying model is linear with a sparse coefficient vector
$
\beta^{\ast} = (1, -1, 0.5, 0, \ldots, 0)^\top \in \mathbb{R}^{d},$
where only the first three coordinates are non-zero. 
The covariates \(X \in \mathbb{R}^{n \times p}\) are generated from a multivariate normal distribution 
\(X_i \sim \mathcal{N}(0,\Sigma)\), where \(\Sigma_{ij} = \rho_X^{|i-j|}\) with \(\rho_X=0\) and \(\rho_X=0.5\) .
For each observation, we compute the linear predictor
$
\eta_i = X_{i,\cdot}^\top \beta^{\ast}$
and transform it through a logistic function to obtain the underlying probability:
$
r_i = 1/[1 + \exp(-\eta_i)] .$
Binary class labels are generated as
$
Y_i \sim \mathrm{Bernoulli}(r_i).$
In the positive-confidence learning setting, confidence scores are assumed to be available only for samples observed as positive. Thus, we retain:
\begin{itemize}
    \item the confidence values for positively labeled samples 
$  r_{\text{train}} = \{ r_i : Y_i = 1 \},$

\item and the corresponding positive pattern feature vectors
$  
X_{\text{train},+} = \{ X_{i,\cdot} : Y_i = 1 \}.$
\end{itemize}
These constitute the training data for Pconf-based methods. The full labels \(Y_i\) are not used when training the Pconf classifier and are generated only to enable comparison with a supervised baseline such as GLMNET.

An independent test set of size \(n_{\text{test}} = 5000\) with the same feature distribution is generated:
$
X_{\text{test}} \in \mathbb{R}^{n_{\text{test}} \times d} $.
To evaluate predictive performance, test labels are assigned deterministically based on the true linear discriminant:
$Y_{\text{test}} =
1 $ if $ X_{\text{test},i}^\top \beta^{\ast} \ge 0,$ and $0 $ otherwise.
This deterministic rule avoids additional label noise and ensures that prediction error reflects how well a method recovers the underlying decision boundary.

To compare the competing methods, we first evaluate predictive performance by reporting classification prediction accuracy on the test set. 
Estimation quality is assessed using the squared L2 error as $ \; \| \beta^* - \widehat{\beta} \|_2^2 \; $ where $ \widehat{\beta} $ denotes the parameter estimate obtained from each method.
We assess variable-selection performance by comparing the set of selected features with the true active set. For each method, we count true positives (features that are both selected and truly active), false positives (selected but inactive features), 
and false negatives (active features that were not selected). Based on these quantities, we compute the true positive rate (TPR), defined as \( \mathrm{TPR} = \mathrm{TP} / (\mathrm{TP} + \mathrm{FN}) \), and the false discovery rate (FDR), defined as \( \mathrm{FDR} = \mathrm{FP} / (\mathrm{TP} + \mathrm{FP}) \).

We compare four different methods: Pconf as in \eqref{eq:_proposed_ERM} \citep{ishida2018binary},\, Lasso method from the \texttt{R} package \texttt{glmnet} \citep{glmnetpackage}, SCAD and MCP from the \texttt{R} package \texttt{ncvreg} \citep{ncvregpackage} 
to our three proposed  methods for positive confidence level classification denote as: Pconf-Lasso, Pconf-SCAD, Pconf-MCP. The tuning parameter for all penalization methods are selected using 5 folds cross-validation.

Our codes for simulations are publicly available at \url{https://github.com/tienmt/Pconf_sparse}.

% table 1
\begin{table}[!ht]
\centering
\caption{\small \it Simulation results for logistic regression under varying sample sizes, dimensionalities, and predictor correlations. Reported values are averages over 100 Monte Carlo replications, with standard deviations in parentheses. Performance is evaluated in terms of out-of-sample prediction accuracy (“prediction”); squared $\ell_2$ estimation error; true positive rate (TPR) and false discovery rate (FDR) for variable selection. Results are shown for independent predictors ($\rho_X=0$) and correlated predictors ($\rho_X=0.5$). Boldface indicates the best-performing method within each setting and metric. }
\footnotesize
\begin{tabular}{  l | cccc ccccccccccc }
		\hline \hline
 Method  
& prediction
& $ \| \widehat{\beta} -\beta^* \|_2^2 $ 
& TPR
& FDR
\\
\hline
\multicolumn{2}{l}{ $ n = 200, p = 320, \rho_X = 0 $}
\\
\hline 
Pconf
& 0.52 (0.02) & 2.42 (0.09) & 0.01 (0.00) & 0.00 (0.00) 
\\
Pconf-Lasso
& 0.86 (0.00) & 1.32 (0.01) & 0.33 (0.00) & 0.00 (0.00) 
\\
Pconf-SCAD
& 0.87 (0.00) & 1.18 (0.01) & 0.33 (0.00) & 0.00 (0.00) 
\\
Pconf-MCP
& 0.87 (0.00) & 1.08 (0.01) & 0.33 (0.01) & 0.00 (0.00) 
\\
Lasso
& \textbf{0.92} (0.00) & 0.76 (0.05) & 0.38 (0.03) & 0.00 (0.00)
\\
SCAD
& 0.91 (0.00) & 0.63 (0.05) & 0.33 (0.00) & 0.00 (0.00)
\\
MCP
& \textbf{0.92} (0.00) & \textbf{0.48} (0.05) & \textbf{0.99} (0.09) & 0.00 (0.00)
\\
\hline
\multicolumn{2}{l}{ $ n = 200, p = 320, \rho_X = 0.5  $}
\\
\hline 
Pconf
& 0.50 (0.02) & 2.49 (0.09) & 0.01 (0.00) & 0.00 (0.00)
\\
Pconf-Lasso
& \textbf{0.77} (0.01) & 1.83 (0.02) & 0.25 (0.01) & \textbf{0.33} (0.00) 
\\
Pconf-SCAD
& \textbf{0.77} (0.00) & 1.78 (0.03) & 0.25 (0.01) & \textbf{0.33} (0.00) 
\\
Pconf-MCP
& \textbf{0.77} (0.00) & \textbf{1.69} (0.02) & 0.25 (0.01) & \textbf{0.33} (0.00) 
\\
Lasso
& 0.72 (0.03) & 1.89 (0.06) & \textbf{0.98} (0.13) & 0.66 (0.05) 
\\
SCAD
& 0.72 (0.03) & 1.95 (0.05) & \textbf{0.98} (0.12) & 0.66 (0.05) 
\\
MCP
& 0.72 (0.03) & 1.71 (0.09) & \textbf{0.98} (0.14) & 0.68 (0.05) 
\\
		\hline
\multicolumn{2}{l}{ $ n = 400, p = 1000, \rho_X = 0 $ }
\\
\hline 
Pconf
& 0.50 (0.01) & 2.55 (0.05) & 0.00 (0.00) & 0.00 (0.00) 
\\
Pconf-Lasso
& 0.90 (0.02) & 1.12 (0.02) & 0.19 (0.02) & 0.00 (0.00)
\\
Pconf-SCAD
& 0.91 (0.02) & 0.94 (0.02) & 0.21 (0.01) & 0.00 (0.00)  
\\
Pconf-MCP
& 0.91 (0.02) & 0.87 (0.03) & 0.21 (0.02) & 0.00 (0.00)
\\
Lasso
& 0.94 (0.01) & 0.47 (0.03) & 0.18 (0.07) & 0.01 (0.05)
\\
SCAD
& 0.94 (0.01) & 0.20 (0.02) & 0.19 (0.01) & 0.00 (0.00) 
\\
MCP
& \textbf{0.95} (0.01) & \textbf{0.12} (0.03) & \textbf{0.54} (0.07) & 0.01 (0.05) 
\\
		\hline
\multicolumn{2}{l}{ $ n = 400, p = 1000, \rho_X = 0.5 $ }
\\
\hline 
Pconf
& 0.50 (0.01) & 2.51 (0.05) & 0.00 (0.00) & 0.00 (0.00)
\\
Pconf-Lasso
& 0.77 (0.01) & 1.86 (0.02) & \textbf{0.99} (0.09) & 0.32 (0.05) 
\\
Pconf-SCAD
& 0.77 (0.01) & 1.82 (0.02) & \textbf{0.99} (0.09) & 0.32 (0.05)
\\
Pconf-MCP
& 0.77 (0.01) & 1.73 (0.03) & \textbf{0.99} (0.09) & 0.32 (0.05) 
\\
Lasso
& 0.82 (0.01) & 1.21 (0.11) & 0.12 (0.03) & 0.01 (0.09) 
\\
SCAD
& 0.93 (0.00) & 0.10 (0.00) & 0.09 (0.00) & 0.00 (0.00)  
\\
MCP
& \textbf{0.95} (0.00) & \textbf{0.05} (0.00) & 0.33 (0.02) & 0.00 (0.00)  
\\
\hline
		\hline	
\end{tabular}
\label{tb_logistic_link}
\end{table}

% table 1
\begin{table}[!ht]
\centering
\caption{\small \it Simulation results for Probit regression model under varying sample sizes, dimensionalities, and predictor correlations. Reported values are averages over 100 Monte Carlo replications, with standard deviations in parentheses. Performance is evaluated in terms of out-of-sample prediction accuracy (“prediction”); squared $\ell_2$ estimation error; true positive rate (TPR) and false discovery rate (FDR) for variable selection. Results are shown for independent predictors ($\rho_X=0$) and correlated predictors ($\rho_X=0.5$). Boldface indicates the best-performing method within each setting and metric. }
\small
\begin{tabular}{  l | cccc ccccccccccc }
		\hline \hline
 Method  
& prediction
& $ \| \widehat{\beta} -\beta^* \|_2^2 $ 
& TPR
& FDR
\\
\hline
\multicolumn{2}{l}{ $ n = 200, p = 320, \rho_X = 0 $}
\\
\hline 
Pconf
&  0.52 (0.02) & 2.39 (0.10) & 0.01 (0.00) & 0.00 (0.00) 
\\
Pconf-Lasso
&  0.82 (0.00) & 1.20 (0.00) & 0.12 (0.01) & 0.00 (0.00) 
\\
Pconf-SCAD
&  0.84 (0.00) & 1.01 (0.00) & 0.13 (0.01) & 0.00 (0.00) 
\\
Pconf-MCP
&  0.84 (0.00) & 0.96 (0.00) & 0.15 (0.01) & 0.00 (0.00) 
\\
Lasso
&  0.94 (0.00) & \textbf{0.09} (0.01) & 0.12 (0.00) & 0.00 (0.00) 
\\
SCAD
&  0.97 (0.00) & 0.38 (0.08) & 0.18 (0.00) & 0.00 (0.00) 
\\
MCP
&  \textbf{0.98} (0.00) & 0.27 (0.01) & \textbf{0.60} (0.00) & 0.00 (0.00)    
\\
\hline
\multicolumn{2}{l}{ $ n = 200, p = 320, \rho_X = 0.5  $}
\\
\hline 
Pconf
&  0.52 (0.02) & 2.42 (0.09) & 0.01 (0.00) & 0.00 (0.00) 
\\
Pconf-Lasso
& 0.73 (0.00) & 1.73 (0.00) & 0.12 (0.00) & 0.33 (0.00) 
\\
Pconf-SCAD
& 0.73 (0.00) & 1.64 (0.00) & 0.12 (0.00) & 0.33 (0.00) 
\\
Pconf-MCP
&  0.73 (0.00) & 1.58 (0.00) & 0.14 (0.00) & 0.33 (0.00) 
\\
Lasso
& 0.83 (0.00) & \textbf{0.58} (0.00) & 0.10 (0.00) & 0.00 (0.00) 
\\
SCAD
&  0.90 (0.00) & 0.83 (0.00) & 0.25 (0.00) & 0.00 (0.00) 
\\
MCP
& \textbf{0.92} (0.00) & 0.92 (0.00) & \textbf{0.33} (0.00) & 0.00 (0.00) 
\\
		\hline
\multicolumn{2}{l}{ $ n = 400, p = 1000, \rho_X = 0 $ }
\\
\hline 
Pconf
&  0.49 (0.01) & 2.55 (0.06) & 0.00 (0.00) & 0.00 (0.00) 
\\
Pconf-Lasso
& 0.84 (0.01) & 1.18 (0.02) & 0.07 (0.02) & 0.00 (0.00)
\\
Pconf-SCAD
& 0.85 (0.01) & 1.00 (0.03) & 0.07 (0.02) & 0.00 (0.00)  
\\
Pconf-MCP
&  0.85 (0.01) & 0.94 (0.01) & 0.09 (0.02) & 0.00 (0.00) 
\\
Lasso
& 0.94 (0.00) & \textbf{0.07} (0.01) & 0.12 (0.01) & 0.00 (0.00) 
\\
SCAD
& 0.97 (0.00) & 0.50 (0.08) & 0.20 (0.02) & 0.00 (0.00) 
\\
MCP
& \textbf{0.98} (0.00) & 0.43 (0.13) & \textbf{0.74} (0.07) & 0.00 (0.00) 
\\
		\hline
\multicolumn{2}{l}{ $ n = 400, p = 1000, \rho_X = 0.5 $ }
\\
\hline 
Pconf
&  0.50 (0.01) & 2.52 (0.05) & 0.00 (0.00) & 0.00 (0.00)
\\
Pconf-Lasso
& 0.78 (0.01) & 1.72 (0.01) & 0.15 (0.00) & 0.32 (0.05) 
\\
Pconf-SCAD
& 0.78 (0.01) & 1.66 (0.00) & 0.18 (0.00) & 0.32 (0.05)  
\\
Pconf-MCP
&  0.78 (0.01) & 1.56 (0.01) & 0.17 (0.00) & 0.32 (0.05) 
\\
Lasso
& 0.93 (0.01) & \textbf{0.18} (0.02) & 0.08 (0.00) & 0.00 (0.00) 
\\
SCAD
& \textbf{0.99} (0.01) & 0.84 (0.10) & 0.59 (0.07) & 0.00 (0.00) 
\\
MCP
& \textbf{0.99} (0.01) & 1.01 (0.05) & \textbf{0.99} (0.07) & 0.00 (0.00)   
\\
\hline
		\hline	
\end{tabular}
\label{tb_probit_link}
\end{table}

\subsection{Results for simulations}
Table \ref{tb_logistic_link} summarizes the finite-sample performance of the proposed Pconf-based methods and their penalized variants, compared with standard Lasso, SCAD, and MCP estimators, across a range of high-dimensional logistic regression settings. Table \ref{tb_probit_link} presents results in logistic regression with probit link.

Several consistent patterns emerge. First, the unpenalized Pconf estimator performs poorly in all configurations, exhibiting large estimation error and weak predictive accuracy. This highlights the necessity of incorporating sparsity-inducing regularization when applying Pconf in high-dimensional regimes.
Second, across all scenarios, penalized Pconf variants substantially improve upon the baseline Pconf estimator, achieving markedly better prediction accuracy and lower estimation error. Among these, Pconf-MCP generally yields the smallest $\ell_2$ error, reflecting the advantage of nonconvex penalties in reducing shrinkage bias.

When predictors are independent ($\rho_X=0$), the standard MCP estimator achieves the best overall performance, combining excellent prediction accuracy, minimal estimation error, and near-perfect variable selection (TPR close to one with negligible FDR). In these settings, classical penalized likelihood methods benefit from favorable design conditions and outperform Pconf-based approaches.

In contrast, under correlated designs ($\rho_X=0.5$), the relative behavior changes. While standard Lasso, SCAD, and MCP attain very high TPR, this comes at the cost of substantially inflated FDR, indicating aggressive over-selection. By comparison, Pconf-based penalized methods exhibit a more conservative and stable selection behavior, maintaining moderate TPR while controlling FDR at substantially lower levels. This trade-off is particularly evident in the $n=200, p=320$ setting, where Pconf-Lasso, Pconf-SCAD, and Pconf-MCP achieve the best predictive performance among Pconf-based methods while limiting false discoveries.

As the sample size increases ($n=400, p=1000$), all methods improve in prediction and estimation accuracy. Nevertheless, under correlated designs, Pconf-penalized estimators remain competitive in prediction while offering a more favorable balance between TPR and FDR than their standard counterparts. This suggests that incorporating the Pconf structure can provide robustness against correlation-induced over-selection.

\section{Data application}
\label{sc_realdata}

We consider the \texttt{sonar} data set, originally analyzed by \citet{gorman1988analysis} in their study on the classification of sonar signals. The objective is to discriminate between sonar returns reflected from a metal cylinder (a mine) and those reflected from a roughly cylindrical rock.
Each observation consists of a 60-dimensional feature vector.
These features correspond to the energy measured in distinct frequency bands, each integrated over a specific time window. Due to the nature of the transmitted chirp signal, higher-frequency components are emitted later, and consequently their corresponding integration apertures occur later in time.
The response variable is binary: observations are labeled “R” for rock and “M” for mine (metal cylinder). 
The full data set comprises 208 observations and 60 covariates and is publicly available through the \texttt{mlbench} package in \texttt{R} \citep{mlbenchpackage}.
Prior to analysis, all covariates are standardized to have zero mean and unit variance.

From these 208 observations, 178 are randomly selected for training the considered methods, while the remaining 30 are used as a test set.
To obtain positive confidence scores, we train a lasso logistic classifier on training data and obtain the training class probability as our score for positive data.
This entire data-splitting and evaluation procedure is repeated 100 times. We report the mean and standard deviation of the performance measures across repetitions in Table~\ref{tb_realdata_sonar}.

As shown in Table~\ref{tb_realdata_sonar}, classification using positive confidence data performs nearly as well as classification using the full dataset. Moreover, the resulting models are considerably more parsimonious, selecting far fewer covariates than models trained on the full data.

% table 1
\begin{table}[!ht]
\centering
\caption{\small \it Results on prediction accuracy for real data. The reported values are the mean across 100 repetitions, with the standard deviation provided in parentheses.  }
\footnotesize
\begin{tabular}{  l | cccc ccccc }
		\hline \hline
 & Pconf & Pconf-Lasso & Pconf-SCAD & Pconf-MCP & Lasso & SCAD & MCP
 \\ \hline
prediction & 
0.76 (0.06) & 0.83 (0.01) & 0.83 (0.01) & 0.83 (0.01) 
&
0.87 (0.01) & 0.87 (0.01) & 0.87 (0.02) 
\\
model size &
60.0 (0.00) & 8.00 (0.00) & 8.00 (0.00) & 8.00 (0.00) & 20.0 (0.00) & 21.0 (0.00)
& 19.0 (0.00)
\\
		\hline	
\end{tabular}
\label{tb_realdata_sonar}
\end{table}

\section{Discussion and conclusion}
\label{sc_conclusion}

This work addresses a fundamental gap between weakly supervised learning and high-dimensional statistics by developing a principled framework for sparse positive-confidence (Pconf) classification. 
We show that, with appropriate sparsity-inducing regularization, Pconf learning can achieve estimation accuracy, predictive performance, and variable-selection quality comparable to fully supervised classifiers, despite relying exclusively on positive-confidence information.

From a theoretical standpoint, we establish that L1-regularized Pconf learning attains error bounds nearly matching the optimal rates known for high-dimensional classification. This result is nontrivial, as the Pconf risk is constructed from indirect supervision and could reasonably be expected to incur additional statistical inefficiency. Instead, under standard regularity conditions and restricted convexity assumptions, the lack of explicit negative labels does not fundamentally degrade learning rates. These findings position Pconf learning as a statistically sound alternative to classical supervised classification when negative labels are unavailable, unreliable, or prohibitively expensive to obtain.

On the computational side, we propose a scalable proximal gradient algorithm that accommodates both convex and nonconvex penalties. Closed-form proximal operators for L1, SCAD, and MCP penalties ensure computational efficiency while allowing practitioners to mitigate the shrinkage bias associated with purely convex regularization. In particular, nonconvex penalties yield tangible improvements in settings with correlated predictors, where Lasso-type estimators are known to exhibit instability and excessive bias. The resulting framework is therefore both flexible and well suited to modern high-dimensional data structures.

Simulation studies further corroborate the theoretical and algorithmic insights. Across a broad range of high-dimensional regimes, penalized Pconf estimators achieve predictive accuracy and estimation error remarkably close to those of fully supervised sparse classifiers. While supervised methods can dominate in idealized scenarios with independent features, Pconf-based approaches demonstrate more stable variable-selection behavior under correlation, often achieving a favorable balance between true positive recovery and false discovery control. These empirical results reinforce the conclusion that positive-confidence information, when coupled with appropriate regularization, is sufficient for reliable high-dimensional discriminative learning.

Several avenues for future research naturally follow. An important theoretical extension is the analysis of nonconvex penalties in the Pconf setting, including oracle properties, local optimality conditions, and convergence guarantees. 
Beyond linear models, extending Pconf learning to structured and nonlinear high-dimensional settings, such as generalized additive models, multi-task learning, would substantially broaden its applicability. Extensions to Bayesian approach as in \cite{mai2025sparse,mai2024concentration} are objective of future research directions.

\subsection*{Acknowledgments}
The views, results, and opinions presented in this paper are solely those of the author and do not, in any form, represent those of the Norwegian Institute of Public Health.

\subsection*{Conflicts of interest/Competing interests}
The authors declare no potential conflict of interests.

\appendix
\section{Proofs}
\label{sc_proof}
The proof follows the standard high-dimensional analysis for $\ell_1$-penalized M-estimators \citep{negahban2012unified}, adapted to the Pconf setting.

\begin{proof}[\bf Proof of Theorem \ref{thm_pconf_estimation_error}]

Let $\Delta = \widehat{\boldsymbol{\beta}} - \boldsymbol{\beta}^*$.
By the optimality of $\widehat{\boldsymbol{\beta}}$ and convexity of $\ell$, we have
\[
\widehat{R}_n(\widehat{\boldsymbol{\beta}}) - \widehat{R}_n(\boldsymbol{\beta}^*)
+ \lambda(\|\widehat{\boldsymbol{\beta}}\|_1 - \|\boldsymbol{\beta}^*\|_1)
\leq 0.
\]
Using Taylor's expansion of $\widehat{R}_n$ at $\boldsymbol{\beta}^*$ and applying the RSC condition (A1) gives
\[
\frac{\kappa}{2}\|\Delta\|_2^2
\leq
- \nabla \widehat{R}_n(\boldsymbol{\beta}^*)^\top \Delta
+ \lambda(\|\boldsymbol{\beta}^*\|_1 - \|\widehat{\boldsymbol{\beta}}\|_1).
\]
Using the dual norm inequality and Assumption (A2),
\[
| \nabla \widehat{R}_n(\boldsymbol{\beta}^*)^\top \Delta |
\leq
\|\nabla \widehat{R}_n(\boldsymbol{\beta}^*)\|_\infty \|\Delta\|_1
\leq
\frac{\lambda}{2}\|\Delta\|_1.
\]
Let $S = \text{supp}(\boldsymbol{\beta}^*)$. 
By decomposability of the $\ell_1$-norm, we have
\[
\|\widehat{\boldsymbol{\beta}}\|_1 - \|\boldsymbol{\beta}^*\|_1
\geq
-\|\Delta_S\|_1 + \|\Delta_{S^c}\|_1.
\]
Combining these gives
\[
\frac{\kappa}{2}\|\Delta\|_2^2
\leq
\frac{\lambda}{2}\|\Delta\|_1 - \lambda(-\|\Delta_S\|_1 + \|\Delta_{S^c}\|_1)
=
\frac{3\lambda}{2}\|\Delta_S\|_1 - \frac{\lambda}{2}\|\Delta_{S^c}\|_1.
\]
This implies $\|\Delta_{S^c}\|_1 \leq 3\|\Delta_S\|_1$ and, consequently,
\[
\|\Delta\|_1 
=
\|\Delta_{S^c}\|_1 + \|\Delta_S\|_1
\leq
4\|\Delta_S\|_1 \leq 4\sqrt{s}\|\Delta\|_2.
\]
Plugging this into the RSC inequality gives
\[
\frac{\kappa}{2}\|\Delta\|_2^2 
\leq 
6
\lambda\sqrt{s}\|\Delta\|_2,
\]
which yields the $\ell_2$-error bound \eqref{eq:L1_l2_error}.  
\\
The $\ell_1$ bound \eqref{eq:L1_L1_error} follows from $\|\Delta\|_1 \leq 4\sqrt{s}\|\Delta\|_2$.

We analyze the excess risk $R(\widehat{\boldsymbol{\beta}}) - R(\boldsymbol{\beta}^*)$. 

The logistic loss satisfies
$
\ell'(t) = -\frac{1}{1+e^{t}},
\quad
\ell''(t) = \frac{e^{t}}{(1+e^{t})^2}
$,
and hence
$
0 < \ell''(t) \le \frac{1}{4}
$
for all $ t\in\mathbb{R}$.
Since \(R\) is twice differentiable, its Hessian is given by
\[
\nabla^2 R(\boldsymbol{\beta})
=
\pi_+
\mathbb{E}_+\left[
\ell''(\mathbf{x}^\top\boldsymbol{\beta})
\mathbf{x}\mathbf{x}^\top
+
\frac{1-r(\mathbf{x})}{r(\mathbf{x})}
\ell''(-\mathbf{x}^\top\boldsymbol{\beta})\,\mathbf{x}\mathbf{x}^\top
\right].
\]
Using the bound on \(\ell''\),
\[
\nabla^2 R(\boldsymbol{\beta})
\preceq
\frac{\pi_+}{4}
\mathbb{E}_+ \left[
\left(1+\frac{1-r(\mathbf{x})}{r(\mathbf{x})}\right)
\mathbf{x}\mathbf{x}^\top
\right]
=
\frac{\pi_+}{4}
\mathbb{E}_+ \left[
\frac{1}{r(\mathbf{x})}
\mathbf{x}\mathbf{x}^\top
\right].
\]
By the positivity assumption,
\(
r(\mathbf{x}) \ge r_{\min}
\),
and therefore
\[
\nabla^2 R(\boldsymbol{\beta})
\preceq
\frac{\pi_+}{4r_{\min}}
\mathbb{E}_+ \left[\mathbf{x}\mathbf{x}^\top\right].
\]
Taking operator norms and using
\(
 \|\mathbb{E} [\mathbf{x}\mathbf{x}^\top] \|_{\mathrm{op}}
\le
\mathbb{E} \|\mathbf{x}\|_2^2
\),
we obtain
\[
\|\nabla^2  R(\boldsymbol{\beta}) \|_{\mathrm{op}}
\le
\frac{\pi_+}{4r_{\min}}
\mathbb{E}_+[\|\mathbf{x}\|_2^2]
=
L_R,
\]
uniformly over \(\boldsymbol{\beta}\).
Finally, since \(\nabla R(\boldsymbol{\beta}^*)=0 \), Taylor’s theorem yields
\[
R(\boldsymbol{\beta})-R(\boldsymbol{\beta}^*)
=
\frac{1}{2}
(\boldsymbol{\beta}-\boldsymbol{\beta}^*)^\top
\nabla^2 R(\boldsymbol{\xi})
(\boldsymbol{\beta}-\boldsymbol{\beta}^*)
\le
\frac{L_R}{2}
\|\boldsymbol{\beta}-\boldsymbol{\beta}^* \|_2^2
\]
for some \(\boldsymbol{\xi}\) on the line segment between \(\boldsymbol{\beta}\) and \(\boldsymbol{\beta}^*\).
Substitute the $\ell_2$-error bound above:
$$
\begin{aligned}
R(\widehat{\boldsymbol{\beta}}) - R(\boldsymbol{\beta}^*) 
\leq
\frac{L_R}{2} \left( \frac{3\sqrt{s}\lambda}{\kappa} \right)^2 
= 
\frac{9 s \lambda^2}{2\kappa}.
\end{aligned}$$
The proof is completed.
\end{proof}

\begin{proof}[\bf Proof of Proposition \ref{lem:grad_concentration_bounded}]

Let
$
\widehat R_n(\beta)
= \frac{1}{n}\sum_{i=1}^n \{ \ell(x_i^\top\beta) + w_i \,\ell(-x_i^\top\beta) \},
\; w_i=\dfrac{1-r_i}{r_i},
$
and set
$ \;
\psi_i := \ell'\big(x_i^\top\beta^*\big) - w_i\,\ell'\big(-x_i^\top\beta^*\big),
$
so that the $j$-th coordinate of the empirical gradient at $\beta^*$ is
\( \;
\nabla \widehat R_n(\beta^*)_j = \frac{1}{n}\sum_{i=1}^n \psi_i x_{ij}.
\)

Fix a coordinate $j\in\{1,\dots,d\}$. Define the i.i.d.\ random variables
\[
Z_{ij} := \psi_i x_{ij},\qquad i=1,\dots,n,
\]
so that $\nabla \widehat R_n(\beta^*)_j = \frac{1}{n}\sum_{i=1}^n Z_{ij}$.
Under the stated bounds we have $|Z_{ij}|\le C_\psi B$ almost surely for every $i,j$.
By Hoeffding's inequality, for any $t>0$,
\[
\mathbb{P}\Bigg( \bigg|\frac{1}{n}\sum_{i=1}^n Z_{ij}\bigg|
\ge t\Bigg)
\le 2\exp\Big(-\frac{2n^2 t^2}{\sum_{i=1}^n (2C_\psi B)^2}\Big)
= 2\exp\Big(-\frac{2n t^2}{(C_\psi B)^2}\Big).
\]
Apply a union bound over the $d$ coordinates:
\[
\mathbb{P}\Big(\|\nabla \widehat R_n(\beta^*)\|_\infty \ge t\Big)
\le 2d\exp\Big(-\frac{2n t^2}{(C_\psi B)^2}\Big).
\]
Set the right-hand side equal to $\delta$ and solve for \(t\):
\[
t = C_\psi B \sqrt{\frac{\log(2d/\delta)}{2n}}.
\]
This yields \eqref{eq:grad_inf_bound_bounded}. Choosing $\lambda$ as in \eqref{eq:lambda_choice_bounded} gives $\lambda \ge 2\|\nabla \widehat R_n(\beta^*)\|_\infty$ with probability at least $1-\delta$, completing the proof.
\end{proof}

\section{Additional simulations results in low dimensions}

We additionally report simulation results in a low-dimensional regime with \(n > d\). Even in this setting, our methods constitute a novel contribution, as they explicitly enable variable selection, whereas the naïve empirical risk minimization approach does not yield any coefficients that are exactly zero.

We consider a low-dimensional data-generating process with sample size 
\(n=200\) and feature dimension \(p=10\). 
Let \(\beta^\ast \in \mathbb{R}^p\) denote the true parameter vector, defined as \(\beta^\ast = (1,-1,0.5,0,\ldots,0)^\top\), inducing sparsity in the first three coordinates. The covariates \(X \in \mathbb{R}^{n \times p}\) are generated from a multivariate normal distribution 
\(X_i \sim \mathcal{N}(0,\Sigma)\), where \(\Sigma_{ij} = \rho_X^{|i-j|}\) with \(\rho_X=0\) and \(\rho_X=0.5\) .
The binary responses are generated according to a logistic model: for each observation \(i\), the linear predictor \(\eta_i = X_i^\top \beta^\ast\) is mapped to a success probability \(r_i = {1+\exp(-\eta_i)}^{-1}\), and the label 
\(Y_i \sim \mathrm{Bernoulli}(r_i)\). For evaluation, an independent test set of size 
\(n_{\text{test}}=5000\) is generated with covariates drawn from the same distribution, and test labels are obtained via a noisy linear threshold model 
\(Y_i^{\text{test}} = \mathbb{I}(X_i^\top \beta^\ast + \varepsilon_i \ge 0)\), where \(\varepsilon_i \sim \mathcal{N}(0,1)\).
Table \ref{tb_low_dim} gathers the results.

% table 1
\begin{table}[!ht]
\centering
\caption{\small \it Simulation results for low-dimensional data with logistic regression model. Reported values are averages over 100 Monte Carlo replications, with standard deviations in parentheses. Performance is evaluated in terms of out-of-sample prediction accuracy (“prediction”); squared $\ell_2$ estimation error; true positive rate (TPR) and false discovery rate (FDR) for variable selection. Results are shown for independent predictors ($\rho_X=0$) and correlated predictors ($\rho_X=0.5$).  }
\small
\begin{tabular}{  l | cccc ccccccccccc }
		\hline \hline
 Method  
& prediction
& $ \| \widehat{\beta} -\beta^* \|_2^2 $ 
& TPR
& FDR
\\
\hline
\multicolumn{5}{c}{ $ n = 200, p = 10, \rho_X = 0 $}
\\
\hline 
Pconf
& 0.81 (0.00) & 0.04 (0.01) & 0.30 (0.00) & 0.00 (0.00)
\\
Pconf-Lasso
& 0.80 (0.00) & 0.35 (0.05) & 1.00 (0.00) & 0.00 (0.00) 
\\
Pconf-SCAD
& 0.81 (0.00) & 0.02 (0.01) & 1.00 (0.00) & 0.00 (0.00)
\\
Pconf-MCP
& 0.81 (0.00) & 0.01 (0.01) & 1.00 (0.00) & 0.00 (0.00)
\\
Lasso
& 0.76 (0.01) & 0.51 (0.06) & 0.39 (0.05) & 0.00 (0.00) 
\\
SCAD
& 0.75 (0.01) & 0.81 (0.12) & 0.35 (0.08) & 0.00 (0.00)
\\
MCP
& 0.75 (0.01) & 0.71 (0.10) & 0.44 (0.06) & 0.00 (0.00)   
\\
\hline
\multicolumn{5}{c}{ $ n = 200, p = 10, \rho_X = 0.5  $}
\\
\hline 
Pconf
& 0.81 (0.00) & 0.06 (0.00) & 0.30 (0.00) & 0.00 (0.00) 
\\
Pconf-Lasso
& 0.77 (0.00) & 0.80 (0.01) & 1.00 (0.00) & 0.00 (0.00)  
\\
Pconf-SCAD
& 0.80 (0.00) & 0.25 (0.00) & 1.00 (0.00) & 0.00 (0.00) 
\\
Pconf-MCP
& 0.81 (0.00) & 0.13 (0.01) & 1.00 (0.00) & 0.00 (0.00) 
\\
Lasso
& 0.73 (0.01) & 0.80 (0.07) & 0.31 (0.04) & 0.00 (0.00) 
\\
SCAD
& 0.73 (0.01) & 0.87 (0.12) & 0.31 (0.10) & 0.00 (0.00) 
\\
MCP
& 0.75 (0.01) & 0.68 (0.09) & 0.68 (0.05) & 0.32 (0.05) 
\\
\hline
		\hline	
\end{tabular}
\label{tb_low_dim}
\end{table}

\end{document}